\documentclass[runningheads]{llncs}
\usepackage{graphicx}
\usepackage[dvipsnames]{xcolor}
\usepackage[caption=false]{subfig}
\usepackage{amsmath,amssymb,amsfonts}

\graphicspath{{figs/}}

\usepackage{tikz}
\usepackage{tikzit}

\usetikzlibrary{arrows.meta}

\tikzstyle{default box}=[fill={rgb,255: red,108; green,172; blue,255}, draw=black, shape=rectangle, inner sep=8pt,rounded corners=0.2cm, line width=0.2mm]
\tikzstyle{st box}=[fill={rgb,255: red,138; green,228; blue,110}, draw=black, shape=rectangle, inner sep=8pt,rounded corners=0.2cm, line width=0.2mm]

\tikzstyle{Arrow}=[-stealth, thick]
\tikzstyle{ArrowD}=[-{Triangle[scale=2.5, length=4, width=6]}, line width=2mm, dashed, color=gray]

\begin{document}
\title{RAIL: A modular framework for Reinforcement-learning-based Adversarial Imitation Learning\thanks{This work has taken place in the Learning Agents Research
		Group (LARG) at UT Austin.  LARG research is supported in part by NSF
		(CPS-1739964, IIS-1724157, NRI-1925082), ONR (N00014-18-2243), FLI
		(RFP2-000), ARO (W911NF-19-2-0333), DARPA, Lockheed Martin, GM, and
		Bosch.  Peter Stone serves as the Executive Director of Sony AI
		America and receives financial compensation for this work.  The terms
		of this arrangement have been reviewed and approved by the University
		of Texas at Austin in accordance with its policy on objectivity in
		research.}}
%
%
\author{Eddy Hudson\inst{1}\orcidID{0000-0002-8350-3919} \and
Garrett Warnell\inst{1, 2}\orcidID{0000-0003-3846-8787} \and
Peter Stone\inst{1, 3}\orcidID{0000-0002-6795-420X}}
\authorrunning{E. Hudson et al.}
%
\institute{The University of Texas at Austin,  TX 78712, USA \and
US Army Research Laboratory\\
\and Sony AI}
\maketitle              
\begin{abstract}
	
	While Adversarial Imitation Learning (AIL) algorithms have recently led to state-of-the-art results on various imitation learning benchmarks, it is unclear as to what impact various design decisions have on performance. To this end, we present here an organizing, modular framework called Reinforcement-learning-based Adversarial Imitation Learning (RAIL) that encompasses and generalizes a popular subclass of existing AIL approaches. Using the view espoused by RAIL, we create two new IfO (Imitation from Observation) algorithms, which we term SAIfO: SAC-based Adversarial Imitation from Observation and SILEM (Skeletal Feature Compensation for Imitation Learning with Embodiment Mismatch). We go into greater depth about SILEM in a separate technical report \cite{hudson2021skeletal}. In this paper, we focus on SAIfO, evaluating it on a suite of locomotion tasks from OpenAI Gym, and showing that it outperforms contemporaneous RAIL algorithms that perform IfO.

\end{abstract}
\section{Introduction}

In the past decade, the field of deep learning has been punctuated by revolutionary results in computer vision, natural language processing (NLP), and reinforcement learning (RL). Key to this explosive growth has been an implicit modularization that has allowed researchers to work on improving individual modules independently and in parallel. For example, in computer vision, an approach to tackle the ImageNet challenge \cite{DBLP:journals/corr/RussakovskyDSKSMHKKBBF14} can be said to comprise the following two main modules among a host of other components: an optimization algorithm, and the deep network architecture. Research into optimization algorithms has yielded algorithms such as AdaGrad \cite{duchi2011adaptive} and Adam \cite{DBLP:journals/corr/KingmaB14}, while research into deep network architectures has resulted in architectures such as convolutional networks and ResNet \cite{he2016deep}. Progress in NLP can similarly be characterized by modules corresponding to optimization algorithm and network architecture. Research into network architectures for NLP has been taking place independent of the optimization algorithm used, and has resulted in success stories from seq2seq \cite{DBLP:conf/nips/SutskeverVL14} to the Transformer \cite{DBLP:conf/nips/VaswaniSPUJGKP17}.

Recently, there has been rapid progress on imitation learning (IL), especially with respect to a class of new algorithms termed adversarial imitation learning (AIL). Starting from GAIL \cite{DBLP:conf/nips/HoE16} and GAIfO \cite{DBLP:journals/corr/abs-1807-06158}, to DAC \cite{DBLP:conf/iclr/KostrikovADLT19} and OPOLO \cite{NEURIPS2020_92977ae4}, the sample complexity of AIL approaches have been rapidly declining. However, progress in this space has been ad hoc: AIL research lacks the type of organizing and modular characterization that has led to the steady and substantial improvement enjoyed by the other communities mentioned above.

In this work, we aim to provide a modular characterization for a sub-class of AIL algorithms that employs RL. We term this class of algorithms Reinforcement-learning-based Adversarial Imitation Learning (RAIL). We characterize RAIL techniques using a modular framework, with modules corresponding to the RL backbone and the input format to the discriminator. As evidence that the modularization of RAIL can accelerate progress in it, we explore the resulting design space and discover a new RAIL variant for imitation from observation that we call SAC-based Adversarial Imitation from Observation (SAIfO), where SAC \cite{haarnoja2018soft} is a recently proposed RL algorithm. We evaluate SAIfO on a suite of locomotion tasks from OpenAI Gym \cite{1606.01540}, and show that it outperforms recent IfO algorithms that fall under the RAIL umbrella. 

\section{Background}

The ultimate goal in imitation learning is to learn a controller that solves a sequential decision making problem.
Such problems are typically formulated in the context of a Markov decision process (MDP), i.e., a tuple $\mathcal{M} = <\mathcal{S}, \mathcal{A}, T, R, \gamma>$, where $\mathcal{S}$ denotes an agent's state space, $\mathcal{A}$ denotes the agent's action space, $T: \mathcal{S} \times \mathcal{A} \rightarrow \Delta(\mathcal{S})$ denotes the environment model which maps state-action pairs to a distribution over the agent's next state, $R: \mathcal{S} \times \mathcal{A} \times \mathcal{S} \rightarrow \mathbb{R}$ is a reward function that provides a scalar-valued reward signal for state-action-next-state tuples, and $\gamma \in [0,1]$ is a discount factor that specifies how the agent should weight short- vs. long-term rewards.
Solutions to sequential decision making problems are often specified by reactive policies $\pi: \mathcal{S} \rightarrow \Delta(\mathcal{A})$, which specify agent behavior by providing a mapping from the agent's current state to a distribution over the actions it can take.
Machine learning solutions to problems described by an MDP typically search for policies that can maximize the agent's expected sum of future rewards.

The IL problem is typically formulated using an MDP {\em without} a specified reward function, i.e., $\mathcal{M}\setminus R$.
Instead of reward, the agent is provided with {\em demonstration} trajectories---typically assumed to have been generated by an expert---that specifies the desired behavior, i.e., $\tau_E = (s_0, a_0, s_1, a_1, ...)$. Imitation from observation (IfO) is a sub-problem of IL in which the agent does not have access to the actions taken during the demonstration trajectories, i.e., $\tau_E = (s_0, s_1, ...)$.
Techniques designed to solve the IL problem seek to use observed demonstrations to find policies that an imitating agent can use to imitate the demonstrator.

Adversarial imitation learning (AIL) is a particular way to perform IL that has recently come to the fore (Figure \ref{fig1}). AIL leverages the adversarial training technique popularized by GANs \cite{DBLP:journals/corr/GoodfellowPMXWOCB14}, in that both involve the same min-max game with discriminators and generator networks. The discriminator, $D$, is trained to distinguish between the demonstration trajectories and trajectories generated by the imitator.
In particular, the goal of updating $D$ is to drive $\mathbb{E}_{o\sim\tau_E}[D(o)]$ toward $1$ and $\mathbb{E}_{o\sim\tau}[D(o)]$ toward $0$, where $o$ is a segment of the trajectory, $\tau$ represents trajectories recently generated by the imitator, and $\tau_E$ is a dataset of demonstration trajectories. In the seminal AIL algorithm GAIL \cite{DBLP:conf/nips/HoE16} and more recent AIL algorithms ASAF \cite{DBLP:conf/nips/BardeRJPPN20} and ValueDICE \cite{DBLP:conf/iclr/KostrikovNT20}, $o=(s_t, a_t)$, whereas in GAIfO \cite{DBLP:journals/corr/abs-1807-06158} and OPOLO \cite{NEURIPS2020_92977ae4}, $o=(s_t, s_{t+1})$. The generator, which in AIL algorithms is the imitator's policy $\pi$, is trained to induce behavior that elicits large output from $D$, i.e., to ``fool" $D$ into thinking that the imitator's trajectories came from the demonstrator.
By iteratively updating $D$ and $\pi$ as  described, AIL approaches are able to find imitator policies that successfully mimic the demonstrated behavior. 

\section{RAIL: Reinforcement-learning-based Adversarial Imitation Learning}

In this work, we focus on a specific slice of AIL that we term RAIL: Reinforcement-learning-based Adversarial Imitation Learning. As suggested by the name, we include in RAIL all AIL algorithms that optimize the policy by using an RL algorithm where the reward is given by the discriminator. Thus, GAIL and GAIfO are RAIL algorithms, as are OPOLO and DAC \cite{DBLP:conf/iclr/KostrikovADLT19}. On the other hand, ValueDICE is not a RAIL algorithm since it optimizes the policy by backpropagating directly into the discriminator, and ASAF is not a RAIL algorithm since it optimizes the policy by simply copying over the weight values from the discriminator. 

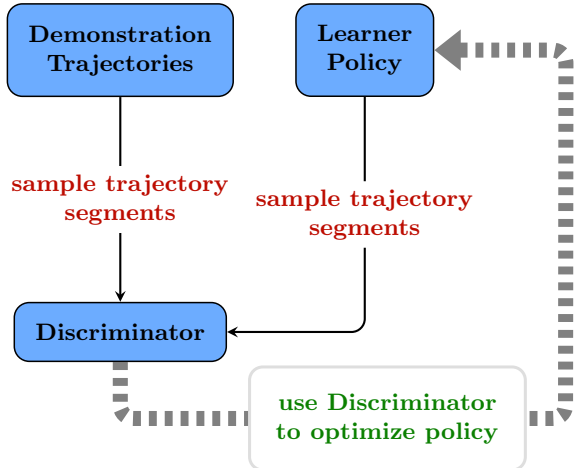
\begin{figure}
	\centering
	\begin{tikzpicture}
	\begin{pgfonlayer}{nodelayer}
		\node [style=default box, align=center] (0) at (-4.25, 3.75) {\textbf{Demonstration} \\ \textbf{Trajectories}};
		\node [style=default box, align=center] (1) at (2.25, 3.75) {\textbf{Learner} \\ \textbf{Policy}};
		\node [style=default box] (2) at (-4.25, -3.75) {\textbf{Discriminator}};
	\end{pgfonlayer}
	\begin{pgfonlayer}{edgelayer}
		\draw [style=Arrow, rounded corners=5pt] (1) |- (2) node[near start, align=center,fill=white, text={rgb,255:red,189;green,12;blue,0}]{\textbf{sample trajectory} \\ \textbf{segments}};
		\draw [style=Arrow] (0) -- (2) node[midway, align=center,fill=white, text={rgb,255:red,189;green,12;blue,0}] {\textbf{sample trajectory} \\ \textbf{segments}};
		\draw [style=ArrowD, rounded corners=5pt] (2) |- node[pos=0.8, align=center,fill=white, draw={rgb,255:red,222;green,222;blue,222}, inner sep=10pt, line width=0.4mm, solid,rounded corners=5pt, text={rgb,255:red,11;green,128;blue,0}] {\textbf{use Discriminator} \\ \textbf{to optimize policy}} ([shift={(90mm,-15mm)}]2.south east) -- ([shift={(35mm,0mm)}]1.east) -- (1);
	\end{pgfonlayer}
\end{tikzpicture}
	\caption{Schematic diagram explaining the work-flow in AIL algorithms} 
	\label{fig1}
\end{figure}

We characterize RAIL techniques using a modular framework consisting of the following two modules (color coded according to the corresponding design decisions in Figure \ref{fig1}):

\begin{itemize}
	\item \textbf{\textcolor{rgb,255:red,11;green,128;blue,0}{the RL backbone}}
	\item \textbf{\textcolor{rgb,255:red,189;green,12;blue,0}{the discriminator's input representation}}
\end{itemize}

The first module is the RL algorithm used by a RAIL algorithm to optimize the learner's policy. Early RAIL algorithms like GAIL or GAIfO, used on-policy RL algorithms such as TRPO \cite{DBLP:conf/icml/SchulmanLAJM15} or PPO \cite{DBLP:journals/corr/SchulmanWDRK17}. More recent RAIL algorithms OPOLO and DAC \cite{DBLP:conf/iclr/KostrikovADLT19} radically reduced their sample complexity by using off-policy RL algorithms such as TD3 \cite{DBLP:conf/icml/FujimotoHM18} or AlgaeDICE \cite{DBLP:journals/corr/abs-1912-02074}.

The second module is the input representation used by the discriminator. This module controls whether the discriminator gets access to state-action pairs $(s_t, a_t)$, state-next state pairs $(s_t, s_{t+1})$, or any arbitrary subsequence of the trajectory. For example, $(s_t, s_{t+3})$. It is worth pointing out here that most RAIL algorithms either use $(s_t, a_t)$, or $(s_t, s_{t+1})$ as the input to the discriminator. This last module that we just introduced is an especially powerful one as it allows an RL researcher to seamlessly create an IfO algorithm out of any RAIL algorithm by changing the input representation to the discriminator. For example, by changing $(s_t, a_t)$ to $(s_t, s_{t+1})$, as was done to create GAIfO from GAIL (Table \ref{tab1}).

\bgroup
\def\arraystretch{1.2}
\begin{table}[]
	\centering
	\caption{Recent RAIL algorithms arranged according to our modular characterization of RAIL. Off policy algorithms are denoted by green-colored names. $T$ is the affine transform that is learned by SILEM.}\label{tab1}
	\begin{tabular}{|c|c|c|}
		\hline
		\textbf{RAIL Algorithm} & \textbf{RL Backbone} & \textbf{Discriminator Input} \\ \hline
		GAIL                    & TRPO \cite{DBLP:conf/icml/SchulmanLAJM15}                & $(s_t, a_t)$                            \\ \hline
		\textcolor{OliveGreen}{DAC}                     & TD3 \cite{DBLP:conf/icml/FujimotoHM18}                 & $(s_t, a_t)$                            \\ \hline
		GAIfO                   & TRPO \cite{DBLP:conf/icml/SchulmanLAJM15}                 & $(s_t, s_{t+1})$                            \\ \hline
		\textcolor{OliveGreen}{OPOLO}                   & AlgaeDICE \cite{DBLP:journals/corr/abs-1912-02074}           & $(s_t, s_{t+1})$                            \\ \hline
		\textcolor{OliveGreen}{DACfO} \cite{NEURIPS2020_92977ae4}                   & TD3 \cite{DBLP:conf/icml/FujimotoHM18}                 & $(s_t, s_{t+1})$                            \\ \hline
		\textcolor{OliveGreen}{SAIfO}                   & SAC \cite{DBLP:conf/icml/HaarnojaZAL18}                 & $(s_t, s_{t+1})$                            \\ \hline
		SILEM \cite{hudson2021skeletal}                    & PPO \cite{DBLP:journals/corr/SchulmanWDRK17}                & $T(s_t, s_{t+1},  s_{t+2}, s_{t+3})$                            \\ \hline
	\end{tabular}
\end{table}
\egroup

Apart from the two modules defined above, RAIL algorithms also typically incorporate miscellaneous tricks to improve performance such as absorbing states in DAC, and regularization with a dynamics model in OPOLO. We disregard them in our modular characterization due to questions that have arisen regarding their value. The authors of the OPOLO paper observe that the regularization only adds marginal improvements in performance \cite{NEURIPS2020_92977ae4}, and our own experience has indicated that absorbing states are not necessary for state of the art performance.

\section{SAIfO}

Given our modular characterization of RAIL algorithms, a natural way to try to obtain a state-of-the-art RAIL algorithm would be to try to maximize the performance of each module independently of one another. Inspired by a recent IfO algorithm, OPOLO \cite{NEURIPS2020_92977ae4}, we develop a state-of-the-art IfO algorithm by maximizing the performance of the RL backbone.  We could also attempt to improve performance by choosing a different form of input to the discriminator that does not include action information, but we leave that to future work. 

We maximize performance of the RL backbone by choosing a state-of-the-art off-policy RL algorithm, Soft Actor Critic (SAC) \cite{DBLP:conf/icml/HaarnojaZAL18}, and we term the resulting RAIL algorithm SAC-based Adversarial Imitation from Observation (SAIfO).

\section{Experiments and Results}

We run experiments on locomotion tasks provided by OpenAI gym in the Mujoco environment. Our implementation of SAIfO is built on the SAC implementation provided by Spinning Up OpenAI \cite{SpinningUp2018}. Every other algorithm we use is based on the implementation provided by the authors of OPOLO \cite{NEURIPS2020_92977ae4}. The appendix contains details on the hyperparameters we use in each algorithm along with the neural network architectures we use.

\begin{figure}[]
	\centering
	\subfloat[Legend\label{fig2c}]{
		\includegraphics[width=0.2\linewidth]{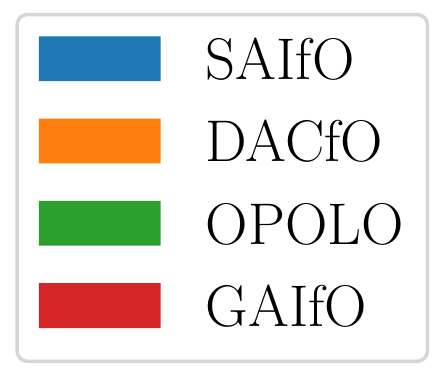}
	} \\
	\subfloat[HalfCheetah-v2\label{fig2b}]{
		\includegraphics[width=0.5\linewidth]{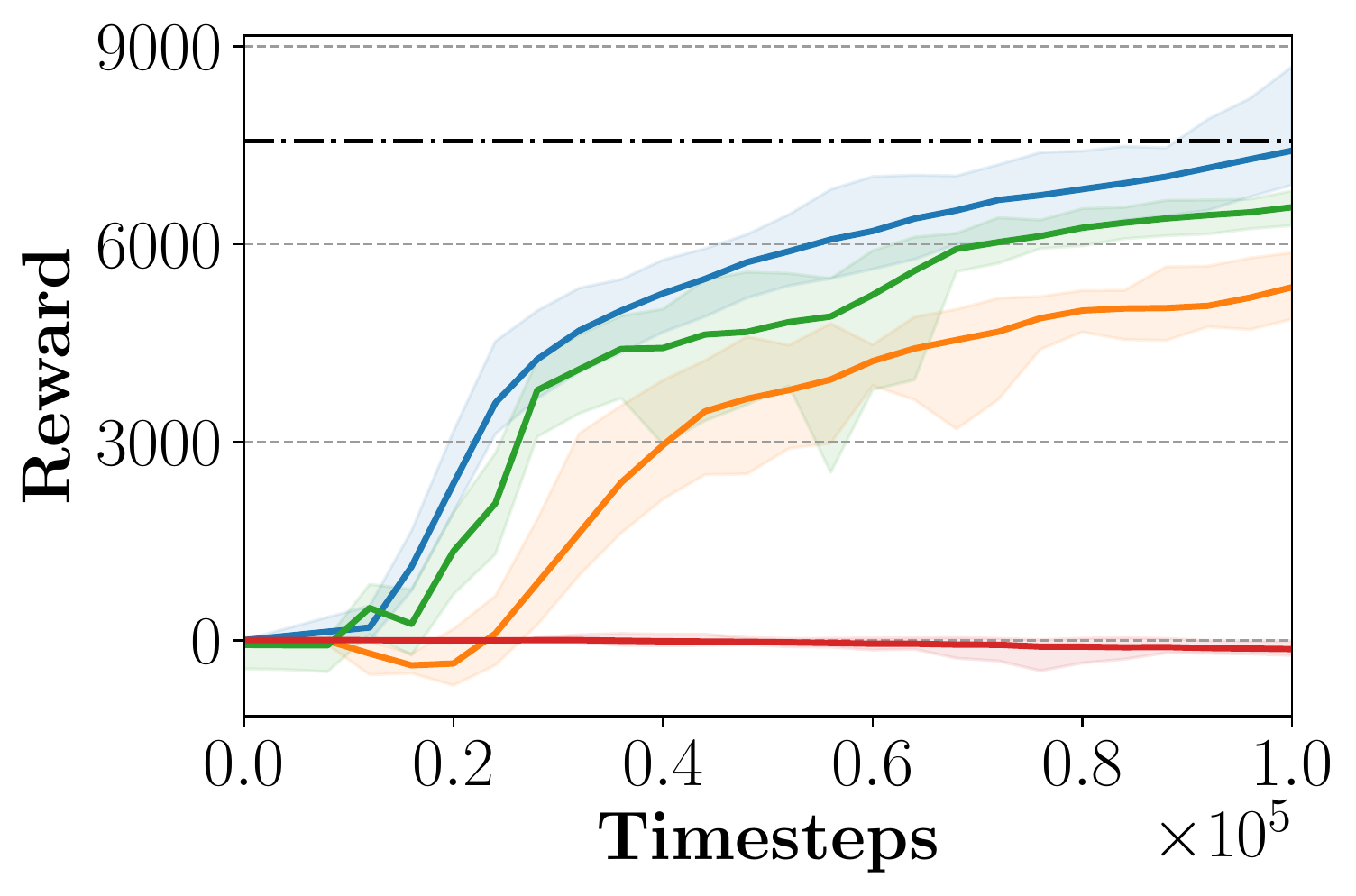}
	}
	\subfloat[Hopper-v2\label{fig2a}]{
		\includegraphics[width=0.5\linewidth]{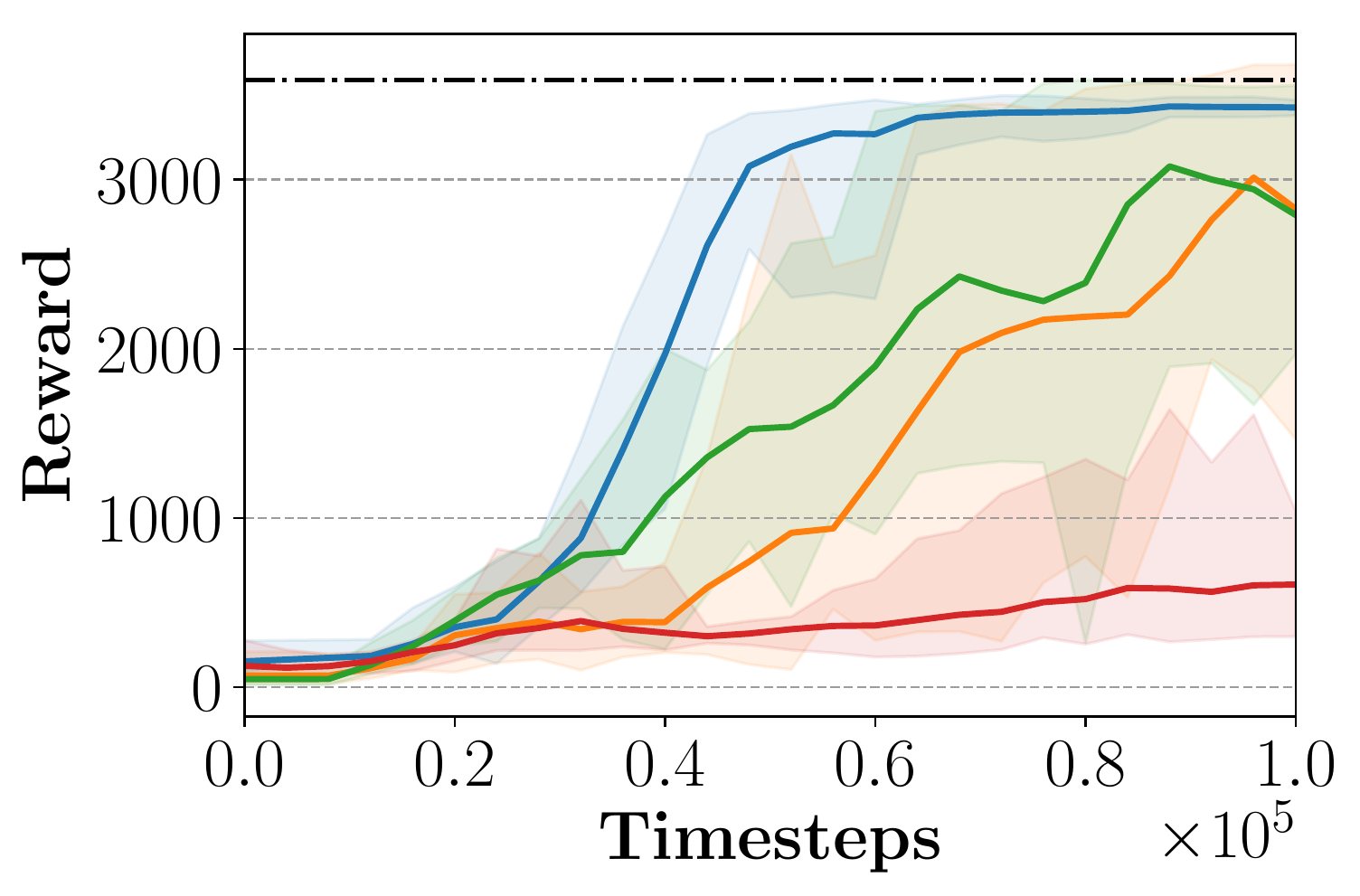}
	}
	\caption{Results comparing SAIfO to RAIL algorithms that perform IfO. The horizontal black lines show expert level performance in each domain. The $x$-axis shows the number of interactions with the environment. Results shown are the mean, minimum and maximum over 10 independent trials.}
	\label{fig2}
\end{figure}

Our main goal with the experiments is to show that our proposed RAIL algorithm, SAIfO, outperforms other RAIL algorithms that perform IfO. We compare SAIfO against GAIfO, OPOLO and DACfO, and find that it indeed does better than the three prior RAIL algorithms (Figure \ref{fig2}). Note that our OPOLO and DACfO results are much better than those reported by Zhu et al. \cite{NEURIPS2020_92977ae4} since we strongly optimized all three of our baselines to provide a fair comparison with SAIfO. Refer to the appendix for specific details on how we optimized their implementations of GAIfO, OPOLO and DACfO.

\section{SILEM}

SAIfO leverages our modular characterization of RAIL to merely accelerate IfO. In SILEM (Skeletal Feature Compensation for Imitation Learning with Embodiment Mismatch) \cite{hudson2021skeletal}, we go a step further and enable IfO to work even in the presence of embodiment mismatch. In SILEM, we process the discriminator's input using a learnable affine transform before inputting it to the discriminator. We train the affine transform to compensate for embodiment mismatch between the learner's trajectories and demonstration trajectories. In Hudson et. al. \cite{hudson2021skeletal}, we evaluate SILEM on a suite of challenging environments involving learning from human demonstrations, and we show that SILEM far outperforms the alternatives. 

\section{Conclusion}

In this work, we introduced a modular characterization of a sub-area in Adversarial Imitation Learning that we term RAIL: Reinforcement-learning-based Adversarial Imitation Learning. We describe the modules apparent in RAIL and show how recent AIL algorithms fit within the modular framework. Using the modularity of the framework as a basis, we develop a new off-policy IfO algorithm (SAIfO) that is faster than prior RAIL algorithms that perform IfO, thus also showing the benefits of the modular characterization.

While developing SAIfO, we only focused on improving performance of the RL backbone. Exploring different formats of the discriminator input is an interesting avenue of future work. For example, rather than using $(s_t, s_{t+1})$ as an input, using $(s_t, s_{t+4})$ or $(s_t, s_{t+1}-s_t)$. We also wish to perform more experiments to figure out why SAC makes a much better RL backbone than the alternatives, TD3 and AlgaeDICE. We suspect that the \emph{stochastic update} afforded by the kernel trick plays an important role \cite{DBLP:conf/icml/HaarnojaZAL18}.

Lastly, we also showcased how our modular characterization of RAIL even allowed us to develop an algorithm capable of performing IfO in the presence of embodiment mismatch. We did this by outlining a recent algorithm we created called SILEM \cite{hudson2021skeletal}.

%
%
%
%
%
\bibliographystyle{splncs04}
\bibliography{example_paper}
\section{Appendix}

\subsection{Ensuring a fair comparison with baselines}

We spent a considerable amount of time optimizing our baselines to ensure a fair comparison. We first added a \emph{warmup} phase to the code provided by the authors of OPOLO where we trained the discriminator and Q function for a configurable number of iterations before using them to train the policy with DACfO or OPOLO. We also applied multiple grid searches to find the best configuration of hyperparameters for our three baselines. We found hyperparameter configurations that resulted in a better performance than the default hyperparameters provided by the OPOLO  authors.

\subsection{Hyperparameters used for each algorithm}

For every experiment, we used a grid search to find the best set of hyper-parameters. Specifically, for every configuration of hyper-parameters in the experiment, we ran 10 independent trials. Within each trial, we trained a policy, measured the average reward obtained by that policy over the last 10 tests (with each test run in intervals of 4000 interactions with the environment), and assigned the resulting number as the score for that particular trial. The score for each configuration of hyper-parameters is the average score over the 10 trials corresponding to that configuration. The best configuration of hyper-parameters is then that which maximizes this score. We usually ran multiple successive grid searches to more strongly optimize performance instead of just running one single grid search for experiment.

\subsubsection{SAIfO} We list below the hyperparameter values we used for our implementation of SAIfO. We only list hyperparameter values that we either introduced or changed from the default value.

\begin{table}[h!]
	\centering
	\caption{Hyperparameter settings for HalfCheetah-v2}\label{Atab1}
	\begin{tabular}{|l|c|}
		\hline
		\multicolumn{1}{|c|}{\textbf{\begin{tabular}[c]{@{}c@{}}Hyperparameter\\ name\end{tabular}}} & \textbf{Value}    \\ \hline
		Discriminator learning rate                                                                  & $6\times 10^{-5}$ \\ \hline
		Entropy coefficient for discriminator                                                        & 0                 \\ \hline
		Batch size (SAC)                                                                             & 400               \\ \hline
		No. steps of warm up for Q function                                                          & 5000              \\ \hline
		No. steps of warm up for Discriminator                                                       & 500               \\ \hline
		Update discriminator every X iterations                                                      & 100               \\ \hline
		SAC alpha                                                                                    & 0.02              \\ \hline
		Gradient penalty for Discriminator                                                           & 0.0006            \\ \hline
		Update policy X times every Y iterations                                                     & 674, 100          \\ \hline
	\end{tabular}
\end{table}

\begin{table}[h!]
	\centering
	\caption{Hyperparameter settings for Hopper-v2}\label{Atab2}
\begin{tabular}{|l|c|}
	\hline
	\multicolumn{1}{|c|}{\textbf{\begin{tabular}[c]{@{}c@{}}Hyperparameter\\ name\end{tabular}}} & \textbf{Value}    \\ \hline
	Discriminator learning rate                                                                  & $1\times 10^{-4}$ \\ \hline
	Entropy coefficient for discriminator                                                        & 0                 \\ \hline
	Batch size (SAC)                                                                             & 400               \\ \hline
	No. steps of warm up for Q function                                                          & 5000              \\ \hline
	No. steps of warm up for Discriminator                                                       & 1000              \\ \hline
	Update discriminator every X iterations                                                      & 10                \\ \hline
	SAC alpha                                                                                    & 0.15              \\ \hline
	Gradient penalty for Discriminator                                                           & 0.002             \\ \hline
	Update policy X times every Y iterations                                                     & 674, 100          \\ \hline
\end{tabular}
\end{table}

\subsubsection{DACfO} We list below the hyperparameter values we used for the implementation of DACfO by the OPOLO authors. We only list hyperparameter values that we either introduced or changed from the default value.

\begin{table}[h!]
	\centering
	\caption{Hyperparameter settings for HalfCheetah-v2}\label{Atab3}
\begin{tabular}{|l|c|}
	\hline
	\multicolumn{1}{|c|}{\textbf{\begin{tabular}[c]{@{}c@{}}Hyperparameter\\ name\end{tabular}}} & \textbf{Value} \\ \hline
	No. steps of warm up for Q function                                                          & 5000           \\ \hline
	No. steps of warm up for Discriminator                                                       & 200            \\ \hline
	Update discriminator every X iterations                                                      & 15             \\ \hline
	Update policy X times every Y iterations                                                     & 7000, 1000     \\ \hline
\end{tabular}
\vspace{33pt}
\end{table}

\begin{table}[h!]
	\centering
	\caption{Hyperparameter settings for Hopper-v2}\label{Atab4}
\begin{tabular}{|l|c|}
	\hline
	\multicolumn{1}{|c|}{\textbf{\begin{tabular}[c]{@{}c@{}}Hyperparameter\\ name\end{tabular}}} & \textbf{Value} \\ \hline
	No. steps of warm up for Q function                                                          & 5000           \\ \hline
	No. steps of warm up for Discriminator                                                       & 100            \\ \hline
	Update discriminator every X iterations                                                      & 300            \\ \hline
	Update policy X times every Y iterations                                                     & 700, 100       \\ \hline
	Batch size                                                                                   & 50             \\ \hline
\end{tabular}
\end{table}

\subsubsection{OPOLO} We list below the hyperparameter values we used for the implementation of OPOLO by the OPOLO authors. We only list hyperparameter values that we either introduced or changed from the default value.

\begin{table}[h!]
	\centering
	\caption{Hyperparameter settings for HalfCheetah-v2}\label{Atab5}
\begin{tabular}{|l|c|}
	\hline
	\multicolumn{1}{|c|}{\textbf{\begin{tabular}[c]{@{}c@{}}Hyperparameter\\ name\end{tabular}}} & \textbf{Value} \\ \hline
	No. steps of warm up for Q function                                                          & 5000           \\ \hline
	No. steps of warm up for Discriminator                                                       & 200            \\ \hline
	Update policy X times every Y iterations                                                     & 2000, 1000     \\ \hline
\end{tabular}
\end{table}

\begin{table}[h!]
	\centering
	\caption{Hyperparameter settings for Hopper-v2}\label{Atab6}
\begin{tabular}{|l|c|}
	\hline
	\multicolumn{1}{|c|}{\textbf{\begin{tabular}[c]{@{}c@{}}Hyperparameter\\ name\end{tabular}}} & \textbf{Value} \\ \hline
	No. steps of warm up for Q function                                                          & 1000           \\ \hline
	No. steps of warm up for Discriminator                                                       & 100            \\ \hline
	Update policy X times every Y iterations                                                     & 700, 100       \\ \hline
	Batch size                                                                                   & 400            \\ \hline
	Update Discriminator every X iterations                                                      & 200            \\ \hline
\end{tabular}
\end{table}

\subsubsection{GAIfO} We list below the hyperparameter values we used for the implementation of GAIfO by the OPOLO authors. We only list hyperparameter values that we either introduced or changed from the default value.

\vspace{33pt}

\begin{table}[h!]
	\centering
	\caption{Hyperparameter settings for HalfCheetah-v2}\label{Atab7}
\begin{tabular}{|l|c|}
	\hline
	\multicolumn{1}{|c|}{\textbf{\begin{tabular}[c]{@{}c@{}}Hyperparameter\\ name\end{tabular}}} & \textbf{Value} \\ \hline
	No. steps of Conjugate gradient descent                                                      & 10             \\ \hline
	No. training steps for value function                                                        & 5              \\ \hline
\end{tabular}
\end{table}

\begin{table}[h!]
	\centering
	\caption{Hyperparameter settings for Hopper-v2}\label{Atab8}
\begin{tabular}{|l|c|}
	\hline
	\multicolumn{1}{|c|}{\textbf{\begin{tabular}[c]{@{}c@{}}Hyperparameter\\ name\end{tabular}}} & \textbf{Value} \\ \hline
	No. steps of Conjugate gradient descent                                                      & 7              \\ \hline
	No. training steps for value function                                                        & 3              \\ \hline
\end{tabular}
\end{table}

\subsection{Neural network architectures}

For every algorithm but SAIfO, we used the default network architecture provided by the authors of OPOLO. For SAIfO, all the deep networks that we used were (multi layer perceptrons) MLPs with two hidden layers and tanh nonlinearities. The discriminator contained 128 units in each hidden layer, while all the other MLPs contained 256 units each.

\end{document}